%% file: root.tex
\documentclass[letterpaper, 10 pt, conference]{ieeeconf}

\IEEEoverridecommandlockouts

\usepackage{etextools}
\usepackage{amssymb}
\usepackage{float}
\usepackage{makeidx}
\usepackage{amsmath}
\usepackage{bbm}
\usepackage{epsfig}
\usepackage{epsf}
\usepackage{psfrag}
\usepackage{verbatim}
\usepackage{color}
\usepackage{multirow}
\usepackage{tabularx}
\usepackage{booktabs}
\usepackage[tight,footnotesize]{subfigure}
\usepackage{array}
\usepackage{soul}
\usepackage{footnote}
\usepackage{cite}
\usepackage{dblfloatfix}

\usepackage{color, colortbl}
\usepackage[colorlinks,bookmarksnumbered,citecolor=orange,urlcolor=orange]{hyperref}
\usepackage{graphicx}
\graphicspath{{Figures}}
\DeclareGraphicsExtensions{.pdf,.png}
\usepackage{bigstrut}
\usepackage[english]{babel}
\graphicspath{{Figures}}
\usepackage{csquotes}

\usepackage[T1]{fontenc}
\usepackage{algorithm, setspace}
\usepackage{algpseudocode}
\usepackage{url}
\usepackage{multirow}
\usepackage{float}
\usepackage{xcolor}
\usepackage{hyperref}
 \hypersetup{
     colorlinks=true,
     linkcolor=orange,
     filecolor=orange,
     citecolor=orange,      
     urlcolor=orange,
     }

\newcommand{\p}[1]{\smallskip \noindent \textbf{{#1}.}}

\newcommand{\fig}[1]{Figure~\ref{fig:#1}}
\usepackage{balance}

\newcommand{\appropto}{\mathrel{\vcenter{
  \offinterlineskip\halign{\hfil$##$\cr
    \propto\cr\noalign{\kern2pt}\sim\cr\noalign{\kern-2pt}}}}}

\title{\LARGE

Aligning Learning with Communication in Shared Autonomy}

\author{Joshua Hoegerman, Shahabedin Sagheb, Benjamin A. Christie, and Dylan P. Losey
\thanks{This work was supported by NSF Grants $\#2222468$ and $\#2129201$.}
\thanks{The authors are members of the Collaborative Robotics Lab (\href{https://collab.me.vt.edu/}{Collab}), Dept. of Mechanical Engineering, Virginia Tech, Blacksburg, VA 24061.
\newline
{e-mail: \texttt{\{jhoegerm, shahab, benc00, losey\}@vt.edu}}}
}

\begin{document}
\maketitle

\begin{abstract}

Assistive robot arms can help humans by partially automating their desired tasks.
Consider an adult with motor impairments controlling an assistive robot arm to eat dinner.
The robot can reduce the number of human inputs --- and how precise those inputs need to be --- by recognizing what the human wants (e.g., a fork) and assisting for that task (e.g., moving towards the fork).
Prior research has largely focused on \textit{learning} the human's task and providing meaningful assistance.
But as the robot learns and assists, we also need to ensure that the human understands the robot's intent (e.g., does the human know the robot is reaching for a fork?).
In this paper, we study the effects of \textit{communicating} learned assistance from the robot back to the human operator.
We do not focus on the specific interfaces used for communication.
Instead, we develop experimental and theoretical models of a) how communication changes the way humans interact with assistive robot arms, and b) how robots can harness these changes to better align with the human's intent.
We first conduct online and in-person user studies where participants operate robots that provide partial assistance, and we measure how the human's inputs change with and without communication.
With communication, we find that humans are more likely to intervene when the robot incorrectly predicts their intent, and more likely to release control when the robot correctly understands their task.
We then use these findings to modify an established robot learning algorithm so that the robot can correctly interpret the human's inputs when communication is present.
Our results from a second in-person user study suggest that this combination of \textit{communication and learning} outperforms assistive systems that isolate either learning or communication.

\end{abstract}


\input{1_intro}
\input{2_related}
\input{3_problem}

\input{4_method}

\input{5_userstudy}

\input{6_conclusion}


\balance
\bibliographystyle{IEEEtran}
\bibliography{bibtex}

\end{document}

%% file: 1_intro.tex
\section{Introduction} \label{sec:intro}

More than $24$ million American adults need external assistance when performing activities of daily living \cite{taylor2018americans}.
Assistive robot arms that \textit{share autonomy} with humans have the potential to help address this challenge \cite{argall2018autonomy, bhattacharjee2020more}.
In these shared autonomy settings the human controls the robot arm using an input device (e.g., a joystick) to indicate their intent, and the robot helps automate tasks on the human's behalf (e.g., picking up foods and feeding them to the operator).

\begin{figure}[t]
    \centering
    \includegraphics[width=0.95\columnwidth]{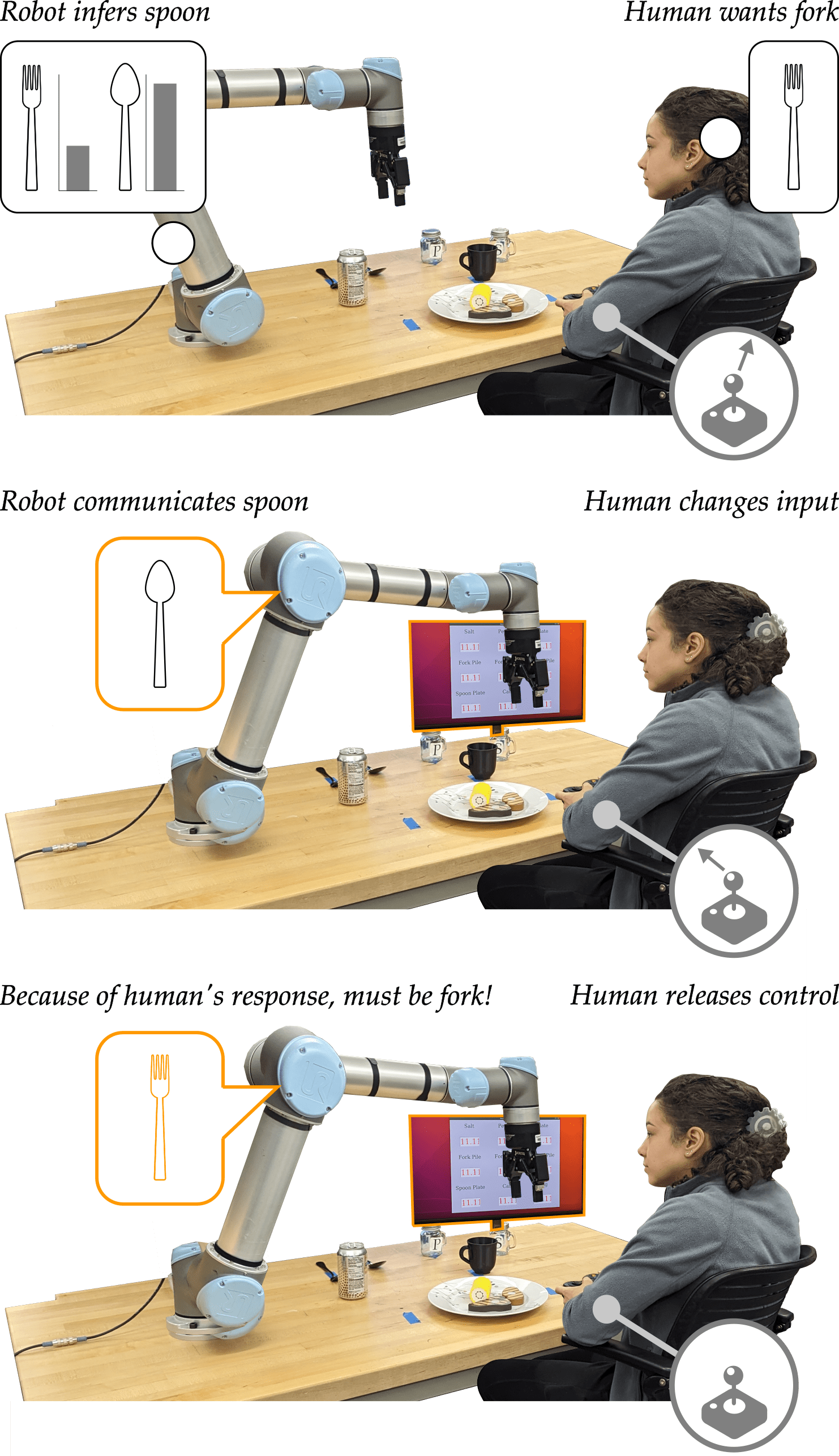}
    \vspace{-0.8em}
    \caption{Human sharing control with an assistive robot arm. (Top) The robot tries to infer the correct task from the human's joystick inputs. (Middle) We show that --- when the robot communicates what it has inferred --- the way humans provide inputs \textit{changes}. (Bottom) If robots are aware of these changes, they can more accurately infer the human's goal.}
    \label{fig:front}
    \vspace{-1.5em}
\end{figure}

To achieve seamless assistance, both the human operator and robot arm must be on the same page.
Consider \fig{front}, where a human is using a robot arm to manipulate kitchen items.
The human wants the robot to pick up a fork, and so the human provides joystick inputs that guide the robot towards that goal.
\textit{For the robot to align with the human}, the robot must \textit{learn} from these inputs to determine the human's intent and partially automate their task.
Here the robot might correctly infer what the human wants (e.g., a fork) and then coordinate its own motions to help reach that goal (e.g., fixing any errors in the human's inputs to precisely pick up the fork).
On the other hand --- \textit{for the human to align with the robot} --- the robot needs to \textit{communicate} its intended assistance back to the user.
Without this communication the human does not know what to expect from the robot: is the robot going to help automate the motion to the fork, or does the robot think the human wants something else entirely?

Existing research on shared autonomy has largely separated learning and communication.
On the one hand, methods such as \cite{javdani2018shared, jain2019probabilistic, dragan2013policy, aronson2018eye, brooks2019balanced} focus on inferring the human's task and partially automating the robot's motion, but do not consider communication back to the human.
On the other hand, approaches like \cite{zolotas2019towards,mullen2021communicating,brooks2020visualization} develop visual and haptic communication interfaces for shared autonomy, but do not modify the robot's learning algorithm.
In this paper, we explore the intersection of learning and communication within shared autonomy settings.
More specifically, we hypothesize that:
\begin{center}\vspace{-0.4em}
\textit{Humans will interact with shared autonomy systems \emph{differently} when those systems communicate their learning.}\vspace{-0.4em}
\end{center}
This is important because --- if humans do provide different inputs in the presence of communication --- then the way the robot interprets and learns from human actions should also be modified.
Accordingly, our paper has two main parts.
First, in Section~\ref{sec:user-study-one} we test our hypothesis and measure how communication can affect the way humans interact with assistive robot arms.
Second, in Sections~\ref{sec:method} and \ref{sec:user-study} we harness the changes caused by communication to modify the robot's learning algorithm.
In practice, this combination of learning and communication enables a) the robot to more seamlessly infer the human's task, and b) the human to more clearly indicate their intent.
Returning to \fig{front}, perhaps the human stops providing inputs because they observe from the robot's feedback that the fork is the robot's most likely goal.
In response, our robot is able to confirm its prediction (i.e., because the human released control the robot must be correct), and complete the task more efficiently.

Overall, we make the following contributions:

\p{Measuring the Effects of Communication}
We consider shared autonomy settings where a human is operating a robot arm, and the robot updates the likelihood of each potential task based on the human's inputs.
For these settings, we perform online and in-person user studies with and without robot communication. We find evidence that humans behave differently in the presence of communication.

\p{Updating the Robot's Learning Rule} Our experimental results suggest that --- when communication is present --- humans are more likely to intervene if the robot has inferred the wrong task, and more likely to relinquish control if the robot is correct. We use these findings to modify the human model of an existing shared autonomy algorithm.

\p{Combining Learning and Communication}
We conduct another in-person user study with three conditions: learning (where the robot does not provide explicit feedback), communication (where the robot communicates its intent but does not adjust its learning rule), and our proposed approach. Our results suggest that the combination of learning and communication increases subjective and objective performance in shared autonomy settings.

%% file: 2_related.tex
\section{Related Works} \label{sec:related}

Below we discuss shared autonomy research that focuses on either learning (i.e., inferring the task and providing assistance) or communication (i.e., visual and haptic interfaces to convey the robot's internal state).

\p{Learning in Shared Autonomy} 
Shared autonomy is a collaborative framework for human-robot interaction where the robot's behavior is a blend of the human's inputs and the robot's autonomous assistance \cite{losey2018review}.
The human's inputs convey the high-level task (e.g., grasping a fork), and the robot's inputs provide fine-grained corrections (e.g., coordinating the motion of the arm to reach that fork).
Prior works develop algorithms to learn both the high-level task and low-level assistance.
For example, in \cite{javdani2018shared, jain2019probabilistic, dragan2013policy, aronson2018eye, brooks2019balanced, fontaine2020quality} the human's desired task is to reach a goal from a discrete set of options, and the robot infers this goal based on the human's inputs.
As the robot becomes more confident in which goal the human wants, it can increasingly provide assistance to automate that task.
Similarly, in \cite{broad2020data, reddy2018shared, schaff2020residual, hagenow2021corrective} the robot builds an estimate of the task's reward function, and overrides any accidental or incorrect human inputs that would result in poor performance (e.g., preventing the human from moving the robot arm into a collision).
Other methods such as \cite{zurek2021situational, cui2023no, he2023learning, losey2022learning, jonnavittula2022sari} learn to assist the human by imitating their previous behaviors. For instance, if the human showed the robot how to pick up a fork in a past interaction, the robot leverages that data to help pick up forks during future interactions.
Overall, each of these works provides a way for the robot to learn from and assist the human.
However, they do not explicitly communicate what the robot has learned --- hence, the user may not know what to expect from the autonomous agent.

\p{Communication in Shared Autonomy} 
Research outside of shared autonomy contexts suggests that communicating robot learning has benefits for both the human and the robot.
From the human's perspective, communication increases the user's acceptance and trust in the system \cite{rosen2019communicating}; from the robot's perspective, communication can result in more effective human teaching and accelerated robot learning \cite{habibian2023review}.
Accordingly, recent works have started to apply communication strategies to shared autonomy \cite{alonso2018system}.
In some scenarios, it is possible for the robot to \textit{implicitly} convey what it has learned by exaggerating its motions \cite{jonnavittula2022communicating}.
However, for the robot to clearly indicate its latent state in everyday settings, \textit{explicit} communication with visual, auditory, or haptic interfaces is often necessary.
In \cite{zolotas2019towards} and \cite{brooks2020visualization} augmented reality headsets show the operator what the robot has learned about their high-level task (e.g., placing visual markers at the most likely goals) and how the robot plans to assist (e.g., displaying the robot's planned trajectory).
Similarly, in \cite{mullen2021communicating} a wearable haptic interface notifies the human when the shared autonomy system is uncertain about their intent.
Our paper will build upon these related works by using explicit communication to convey the robot's inferred task back to the human.
However, instead of focusing on the communication interface itself, we are interested in the effects of this communication on the human operator and assistive agent.

%% file: 3_problem.tex
\section{Effects of Communication \\in Shared Autonomy} \label{sec:problem}

We consider shared autonomy settings where the human and robot collaborate in achieving a common goal. A key aspect of shared autonomy is the ability of the robot to infer the human's goal (i.e., the task they are trying to complete).
If the robot correctly infers the human's goal, it can complete the remaining task without requiring further human input. Alternatively, if the robot's inference is incorrect, the human must keep providing inputs towards their intended goal. However, it can be challenging for humans to determine what goal the robot has inferred without explicit communication. 

In this section, we investigate how explicitly communicating the robot's belief about the human's goal affects their actions. 
We first introduce the policies of the human and the robot collaborator in the absence of communication. Then, we conduct a user study to understand the role of communication in shared-autonomy settings and determine how the users' actions change when communication is introduced. We aim to use these findings to improve the robot's inference of the human's goal and provide better assistance.


\subsection{Shared Autonomy without Communication}\label{sec:no-communication}

We let $s \in \mathcal{S}$ be the environment state which includes the state of the robot, $a_\mathcal{H} \in \mathcal{A}$ and $a_\mathcal{R} \in \mathcal{A}$ be the human's and robot's actions respectively. 
The environment state transitions based on both the human and robot actions.
\begin{equation}
s^{t+1} = f\left(s^t, a_\mathcal{H}, a_\mathcal{R}\right)
\label{eq:state-transition}
\end{equation}
We assume that the human chooses actions to minimize an internal cost-value function $Q^\star$:
\begin{equation}
a_\mathcal{H} \sim \pi_\mathcal{H}^\star\left(\circ \mid s, \theta, Q^\star\right)
\label{eq:human-policy}
\end{equation}
Correspondingly, as the robot is trying to achieve the same goal as the human, it should take actions that minimize the human's cost-value function $Q^\star$. The robot does not directly observe the human's goal or their cost-value function. Instead, the robot selects actions according to an approximation $Q$ of the cost-value function from prior work~\cite{javdani2018shared} where the robot's belief is not directly communicated to the human:
\begin{equation}
a_\mathcal{R} \sim \pi_\mathcal{R}^\star\left(\circ \mid s, b\left(\theta\right), Q\right)
\label{eq:robot-policy}
\end{equation}
where $b(\theta)$ is the robot's \textit{belief} of the human's goal $\theta$. 



We suspect that the human's actions will change in the presence of communication. If the belief communicated by the robot aligns with the human's goal --- will the human continue to provide actions that navigate the robot towards their goal or will they allow the robot to assume full control? On the other hand, if the robot's belief is incorrect --- will the human \textit{exaggerate} their corrective actions because they know that the robot's belief is incorrect? To evaluate how real users respond to robots that communicate their belief, we conducted two user studies in the absence and presence of communication.


\subsection{Shared Autonomy with Communication}\label{sec:user-study-one}

We performed online and in-person user studies to gain insight into the effect of communication on shared autonomy. Participants collaborated with a robot to reach a goal while choosing how much input they think is enough for the robot to learn the task. In half of the interactions, the robot communicated its current belief of the user goal as a percentage using a digital interface. Our results from $25$ online users and $10$ in-person users show that people provide less input when the robot communicates its belief over the user goal. Additionally, the subjective polled results from the in-person study show a significant preference for a system that communicates the robot's intention for its cooperation.

\begin{figure*}[t]
    \centering
    \includegraphics[width=2.0\columnwidth]{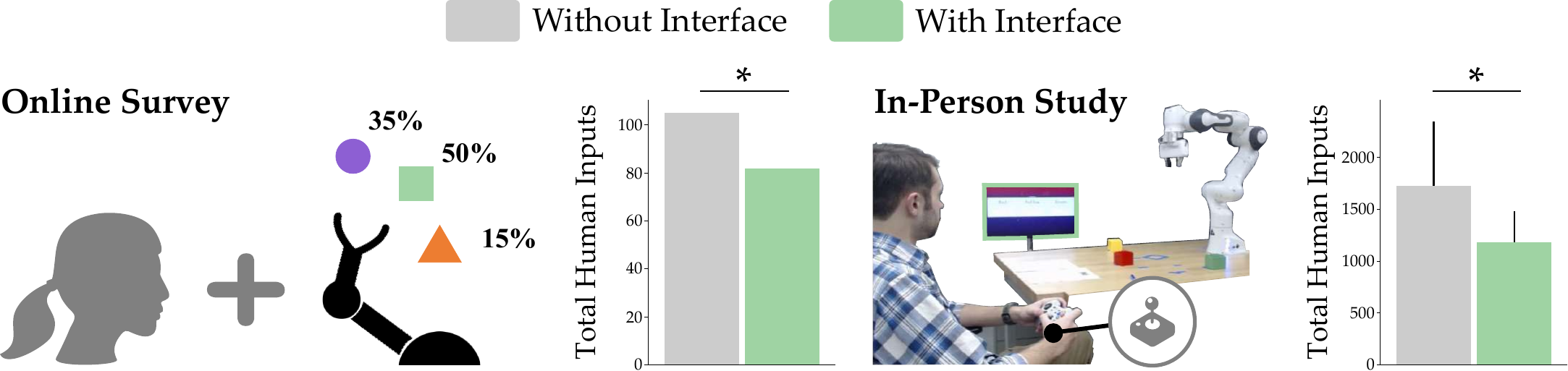}
    \caption{Example settings and results from our user studies in Section~\ref{sec:user-study-one}. Here we explored how communicating the robot's inferred distribution over a discrete set of tasks affected the human's inputs during shared autonomy.
    In all conditions, the robot used the same learning algorithm. 
    (Left) Results from the online survey with and without a communication interface. Humans were more likely to release control to an assistive robot that conveyed its learned distribution over the tasks ($t(24) = 4.271, p < 0.005$).
    (Right) Corresponding results from our in-person study. Here humans required fewer inputs to guide the robot to their goal when the robot communicated its learning ($t(29) = 2.986, p < 0.005$). Overall, these results suggest that humans are more willing to yield control to a communicative system. An asterisk (*) denotes statistical significance. 
    }
    \label{fig:study2}
\end{figure*}

\p{Experimental Setup}
In the online study, participants taught a robot to reach a goal in multiple shared autonomy settings. In each setting, there were three objects with varying colors and the user's goal was to reach the green square (see Figure \ref{fig:study2} (Left)). The position of these objects varied between settings. To simulate the settings we used an animated 2D environment with a top-down view. Online participants first watched the beginning of the robot arm's motion and then selected their choice of input to guide the robot toward the desired goal or to allow the robot to continue on its partially demonstrated path. In the in-person study, users commanded a robot arm using a Logitech F$710$ gamepad to perform a similar task of reaching a green cube. 

We had five different settings for the objects in the online study and three settings in the in-person study. All participants interacted with the robot in each setting twice --- with and without communication. 
In total, participants had six interactions in the in-person study and ten interactions in the online study. Each interaction ended when the robot reached the correct goal. The order of the settings was randomly counter-balanced across all users.

\p{Independent Variables}
For the online study, the users interacted with the robot in each setting across two variations. In one variation, the users had to infer the robot's intended goal through its animated motion (\textbf{Without Interface}). In the other variation, users were provided with the probabilities of the robot's belief over the goals (\textbf{With Interface}) in addition to their observation of the robot's motion.

For the in-person study, the robot used a state-of-the-art shared autonomy algorithm \cite{javdani2018shared} to select its action $a_\mathcal{R}$ in each setting. In half of the interactions, the robot communicated its current belief $b(\theta)$ as percentages using a digital interface (\textbf{With Interface}) and for the other half, the users had to infer the robot's belief from its motion (\textbf{Without Interface}). 

\p{Dependent Variables}
In both studies, we focused on how the user responses change when performing the shared autonomy tasks with and without communication.
For the online study, we recorded whether the human chose to command the robot toward the desired goal or not. For the in-person study, we recorded the time that users spent using the gamepad and other joystick inputs (\textit{Total Human Inputs}) as well as their subjective responses on a 7-point Likert scale for whether they preferred the settings with explicit communication or without the interface.

\p{Participants}
For the online study, we recruited $25$ anonymous participants. We included an instruction and a qualifying question at the beginning of the survey for this study. For the in-person study, we recruited $10$ participants from the Virginia Tech community ($2$ female, ages $23 \pm 9$ years). All participants provided informed consent as per university guidelines (IRB $\#20$-$755$). To assist the participants in becoming familiar with the gamepad and the robot we provided practice time at the beginning of the interaction.

\p{Hypothesis}
We hypothesized that:
\begin{quote}
\p{H1} \textit{When the robot communicates its belief over the goal, users will require less effort in commanding the robot to reach the desired goal.}

\p{H2} \textit{Users will prefer using a shared autonomy system where the robot's belief is communicated.}
\end{quote}

\p{Results}
Our results from the online and in-person user studies are summarized in Figure \ref{fig:study2}. To address \textbf{H1}, we evaluate the level of \textit{effort} that users exhibited through the number of human inputs given through the gamepad. Here, there was a significant difference ($t(24) = 4.271, p < 0.005$) in the requisite human effort to reach the given goals. This result shows that when the robot communicates its belief, the user no longer has to provide the same level of effort for the robot to reach the goal, supporting \textbf{H1}. 

For \textbf{H2}, we turn to our Likert-scale survey. We performed a Paired-Samples T-Test across polled user preferences for communication; these results were significant ($t(9) = 17.676, p < 0.001$). In our in-person user study, participants preferred interacting with a robot that communicated its belief of the human's intent. 




%% file: 4_method.tex
\section{Harnessing Communication \\to Improve Learning} \label{sec:method}

Our results from the first user study (Sec. \ref{sec:user-study-one}) demonstrate that humans behave differently in settings with communication than those without it. In this section, we leverage the human's response to the robot's communication in a novel shared-autonomy formalism. Instead of solely using communication to aid the human's guidance of the robot, we treat the human's feedback to the communication as an indication of the user's confidence in the robot.

We use this idea to present model human policies for both modalities: in the presence and absence of communication. The robot policy uses the appropriate human model to choose assistive actions that minimize the human's modality-specific cost-value function.

\p{Human} The human takes actions that minimize their internal cost-value function $Q$. Following previous works \cite{jonnavittula2022communicating,jain2019probabilistic}, we model the human as a nosily rational agent
according to the Boltzmann distribution: 
\begin{equation}
\pi_\mathcal{H}\left(a_\mathcal{H} \mid s, \theta\right) =
\frac{
\exp\left(\beta \cdot Q\left(s, a_\mathcal{H}, \theta\right)\right)
}{
\int \exp\left(\beta \cdot Q\left(s,a_\mathcal{H}^\prime , \theta, \right)\right) d{a_\mathcal{H}^\prime}
}
\label{eq:human-model}
\end{equation}

Here, $\pi_\mathcal{H}$ is a model of the human's true policy $\pi_\mathcal{H}^\star$. In the Boltzmann rational distribution, $\beta \in [0, \infty)$ is the rationality hyperparameter: as $\beta$ approaches 0, the human is considered to be more irrational; their actions are essentially uniformly distributed. On the other hand, as $\beta$ increases, the human's actions are increasingly optimal (i.e. "rational"). 
The robot does not have access to the human's policy; instead, it assumes an apriori model of the human.
In continuous spaces, Equation \ref{eq:human-model} is intractable. Similar to \cite{jain2019probabilistic}, we tractably estimate the human's policy using the principle of maximum entropy: the probability of a goal decreases exponentially as its cost increases. This yields the following approximation:
\begin{equation}
\pi_\mathcal{H}\left(a_\mathcal{H} \mid s, \theta\right)
 \propto \exp\left(-\beta \cdot Q(s,a_\mathcal{H}, \theta\right))
 \label{eq:propto}
\end{equation}

Firstly, in the absence of communication, we approximate the human's cost-value function as:
\begin{equation}
    \mathcal{Q} (s, a_\mathcal{H}, \theta) = \text{dist}(a_\mathcal{H} + s, \theta) - \text{dist}(s, \theta) + \| a_\mathcal{H} \|
    \label{eq:human-q}
\end{equation}
The first two terms measure the distance by which the human actions move the robot away from the human's goal, while the last term measures the magnitude of the human actions. Formally, Equation \ref{eq:human-q} is minimized when the human takes \textit{low-effort} actions that minimize the distance between the robot and their goal and require the least effort to do so. However, in the absence of communication, humans cannot directly observe whether or not their actions have influenced the robot's belief to a state where they no longer need to provide input actions and thus, have no reliable basis on which to determine when they can minimize their effort.



In the absence of communication, the human must infer the robot's belief by observing the robot's actions.  
However, in many cases there can be uncertainty in determining the robot's goal --- for example, if the spoon and the fork are close to one another, how can the human reliably tell which goal the robot is moving towards? On the other hand, in the presence of communication, the human has a reliable prediction of the robot's future assistive actions given its belief and can respond to this communication \textit{positively} by removing input or \textit{negatively} by continuing to work against the robot. Our key insight is that when the robot's belief is communicated, human inputs can be interpreted as assurance or rebuttal of this communicated belief.

Therefore, we propose that the human's internal cost-value function in the presence of communication can be modeled 
by incorporating the robot's belief into the cost of the human's actions.
\begin{equation}
    \mathcal{Q} (s, a_\mathcal{H}, \theta) = \text{dist}(a_\mathcal{H} + s, \theta) - \text{dist}(s, \theta) + b(\theta) \cdot\| a_\mathcal{H} \|
    \label{eq:human-q-communication}
\end{equation}

In the presence of communication, if the robot's belief is correct, then the human's cost is minimized by providing little effort in agreement with the robot's assistance. If the robot's belief is \textit{incorrect}, then the human will provide inputs that contradict this belief. For example, in the case of ambiguous goals (i.e., the spoon and fork placed close together), with the presence of communication, the human will hold a definite answer for whether the robot is correct. This will result in either further adjustments to correct a misaligned belief or a submission of control seeing that they can minimize their effort by relying on the robot's assistance. 

\p{Robot} The robot updates its belief $b\left(\theta\right)$ based on the observed human actions. Let $P(\theta \mid s, a_\mathcal{H})$ denote the probability that the human is optimizing for the goal $\theta$ given the state $s$ and human action $a_\mathcal{H}$. Using Bayes' theorem, the posterior probability is defined as:
\begin{equation}
P(\theta \mid s, a_\mathcal{H}) \propto P(a_\mathcal{H} \mid s, \theta) \cdot P(\theta)
\label{eq:m8}
\end{equation}
Here, $P(\theta)$ is the prior of the robot's belief over the human's goal and $P(a_\mathcal{H} \mid s, \theta)$ is the likelihood function for the robot's prediction. Note that $P(a_\mathcal{H} \mid s, \theta)$ is equivalent to $\pi_\mathcal{H}^\star$, which we model as $\pi_\mathcal{H}$. 
Similar to Equation \ref{eq:propto}, we use the principle of maximum entropy to derive an equivalent form for Equation \ref{eq:m8}:
\begin{equation}
P(\theta \mid s, a_\mathcal{H}) 
\propto
\exp\left(-\beta \cdot Q\left(s, a_\mathcal{H}, \theta\right)\right) \cdot P(\theta)
\end{equation}

The robot takes actions $a_\mathcal{R}$ that minimize Equation \ref{eq:human-q} in the absence of communication and Equation \ref{eq:human-q-communication} in the presence of communication according to:
\begin{equation}
    a_\mathcal{R} = \sum\limits_{\theta \in \Theta} P\left(\theta \mid s, a_\mathcal{H}\right) \cdot \left(\theta - s\right)
    \label{eq:robot-actions}
\end{equation}
Since the robot's belief may be incorrect, the robot \textit{blends} the human's commanded action with an assistive action:
\begin{equation}
a_\mathcal{B} = \left(1 - \alpha\right) \cdot a_\mathcal{H} + \alpha \cdot a_\mathcal{R}
\label{eq:blended-action}
\end{equation}
The hyperparameter $\alpha \in [0, 1]$ is determined by a threshold according to the human's action such that when the robot displays the correct belief and the human surrenders control, the robot is allowed to take a higher level of control to assist.
For this, we transition alpha from a minimum value in the presence of human action to a maximum value when the robot is in full control.

\begin{equation}
\begin{cases}
\alpha = \alpha + \textit{step}, ~\alpha \le \alpha_\text{max} & 
\text{if} ~~ \|a_\mathcal{H}\| \approx 0
\\
\alpha = \alpha - \textit{step}, ~\alpha \ge \alpha_\text{min} & 
\text{if} ~~ \|a_\mathcal{H}\| \not\approx 0
\end{cases}
\label{eq:cases}
\end{equation}
Here \textit{step} is a hyperparameter chosen by the designer to control the rate at which the robot will increase its assistance proportionally to the number of timesteps that the user has allowed for complete robot assistance.

\begin{figure}[t]
    \centering
    \includegraphics[width=1\columnwidth]{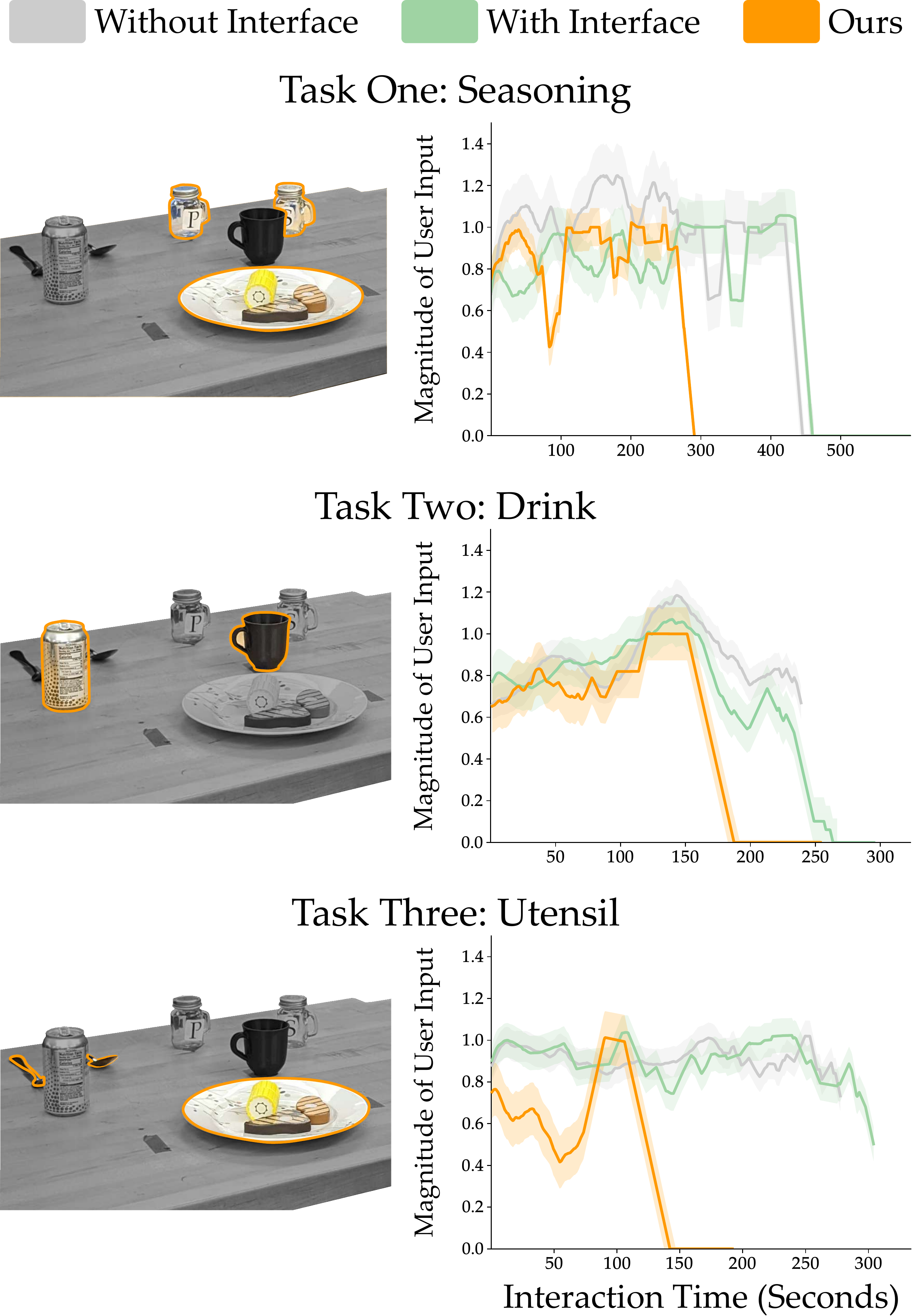}
    \caption{Tasks and user inputs from the user study in Section~\ref{sec:user-study}. (Left) The items the human led the robot to interact within each task. (Right) The magnitude of the human's inputs over time averaged across all users. These results show that users completed the tasks more quickly with \textbf{Ours}, and overall needed fewer inputs to convey their intended goals to the robot.}
    \label{fig:m4}
\end{figure}

\begin{figure*}[t]
    \centering
    \includegraphics[width=2\columnwidth]{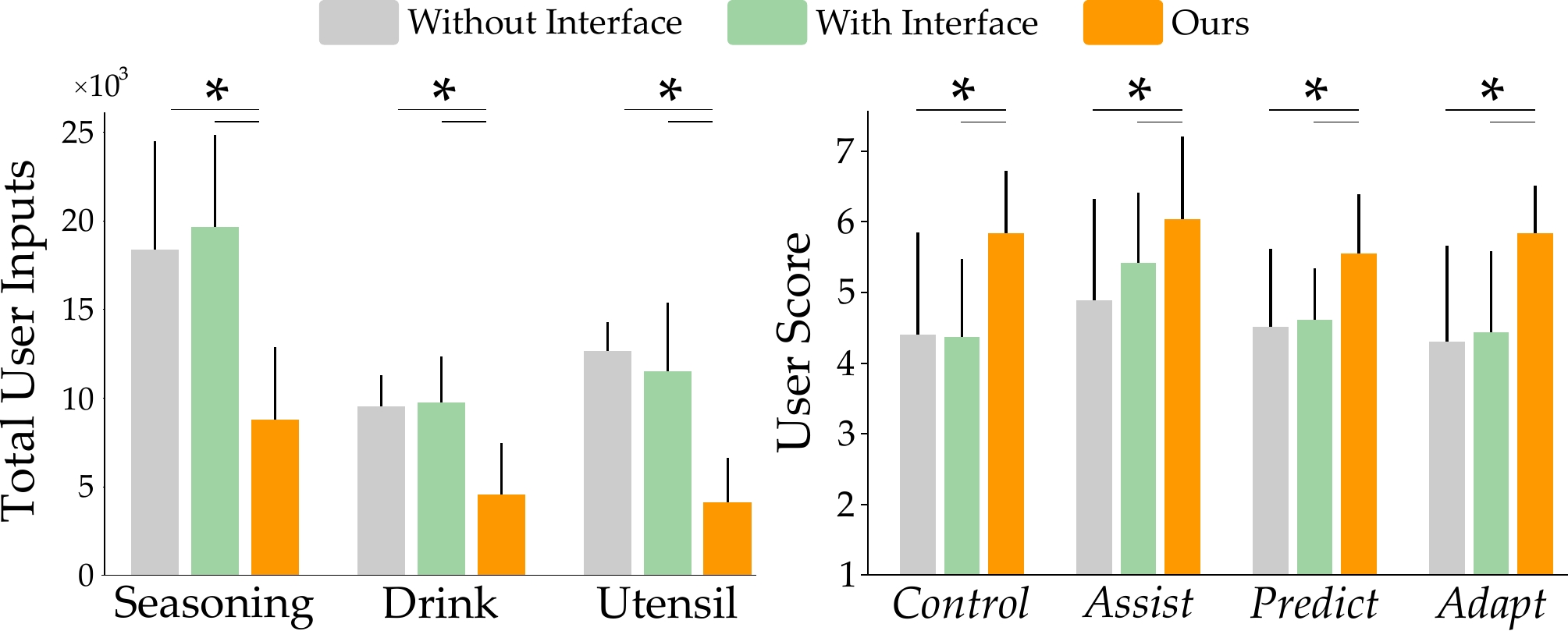}
    \caption{Objective and subjective results from the user study in Section~\ref{sec:user-study}.
    (Left) Total user inputs for \textbf{Seasoning}, \textbf{Drink}, and \textbf{Utensil} tasks. To count the number of inputs, the robot measured whether the human had pressed the joystick every $0.02$ seconds. Across each task, users provided fewer inputs and relied on the robot's assistance more when using \textbf{Ours} ($p < 0.001$, $p < 0.001$, $p < 0.001$). These results support \textbf{H3}: Users spent less effort when using \textbf{Ours}. 
    (Right) Subjective results for the three baselines. Across the four Likert-Scale items, users preferred \textbf{Our} method: they felt that they could easily \textit{control} the system ($p < 0.001$), the robot provided effective \textit{assistance} ($p < 0.005$), the robot better \textit{predicted} their goal ($p < 0.001$), and the robot \textit{adapted} more quickly to their actions ($p < 0.001$).
    }
    \label{fig:us2}
\end{figure*}

Altogether, Equations \ref{eq:human-q-communication}-\ref{eq:cases} form our method for selecting optimal actions in the presence of communication. These equations build upon existing shared autonomy approaches for inferring the human's goal and providing assistance \cite{jain2019probabilistic, dragan2013policy, javdani2018shared}. But we have modified this existing learning framework to explicitly account for communication and the effect communication may have on the human's internal cost function $Q^*$.
Without communication, earlier works such as \cite{javdani2018shared} suggest that the human optimizes for their error and effort as shown in Equation~\ref{eq:human-q}. However, with communication, our experimental results from Section~\ref{sec:user-study-one} indicate that humans are willing to increase their effort if the robot is wrong and release control when the robot is correct.
Using these findings we update our human model for $Q$ in Equation~\ref{eq:human-q-communication}.
Up to this point, our modified learning rule is informed by experiments but has not yet been tested.
Accordingly, in Section~\ref{sec:user-study} we will compare our proposed method for aligning learning with communication against baselines that separately learn and communicate.

%% file: 5_userstudy.tex
\section{Testing the Combination \\of Learning and Communication} \label{sec:user-study}

Lastly, we conduct an in-person user study to evaluate the performance of our proposed method in comparison to the state-of-the-art shared autonomy baseline~\cite{javdani2018shared} with and without a communicative interface. We wish to demonstrate that accounting for the knowledge of the robot's belief in the human's cost function, in addition to communicating the robot's belief, allows the robot to provide better assistance than simply communicating the robot's belief with the baseline shared autonomy approach.

%

\p{Experimental Setup}
Users were instructed to complete three tasks in a more complicated environment than the first in-person user study to highlight the utility of this approach: 
\begin{enumerate}
\item{\textbf{Seasoning:}} Retrieve a salt or pepper shaker, bring it to a plate of food, and then return it to its base position.

\item{\textbf{Drink:}} Go to the can of soda, bring the can of soda to a mug, and return the can to its base position.

\item{\textbf{Utensil:}} Retrieve the spoon or fork and bring it to the relevant side of the plate. 
\end{enumerate}


Participants commanded the robot using a Logitech F$310$ gamepad to complete each of the three tasks using one of three methods: \textbf{Without Interface}, \textbf{With Interface}, and \textbf{Ours}. The order in which participants interacted with these methods was randomized to avoid any proficiency bias. Details of these tasks are illustrated in Figure \ref{fig:m4} (Left).


\p{Independent Variables}
In each task, the robot starts with a uniform prior over the goals which is gradually updated according to the methods discussed in section \ref{sec:method}.
Participants performed each task three times --- using the baseline shared autonomy approach \textbf{Without Interface}, using the same baseline \textbf{With Interface}, and using our method of feedback-enabled shared autonomy - \textbf{Ours} (which combines learning with communication).

\p{Dependent Variables}
We recorded the \textit{Total User Inputs} to measure the amount of effort spent by the users in completing each task. We also recorded subjective \textit{User Scores} through a $7$-point Likert scale survey with four items --- for how easy it was to \textit{Control} the robot, how often they could tell when the robot \textit{Assisted} them, whether the robot was able to \textit{Predict} their goals, and if the robot \textit{Adapted} to their actions. 


\p{Participants}
A total of $12$ participants from the Virginia Tech community took part in this study ($2$ female, ages $28.5 \pm 6.5$ years). Two of the twelve users had not interacted with robots before. Users provided written consent as per university guidelines (IRB $\#20$-$755$).

\p{Hypothesis}
We hypothesized that for this study:
\begin{quote}
\p{H3} \textit{The human will spend less effort in completing the tasks when using} Our \textit{method.}

\p{H4} \textit{Users will provide higher scores on the subjective metrics for} Our \textit{method than the baselines.}
\end{quote}

\p{Results}
The results of our user study are summarized in Figure \ref{fig:us2}. To address \textbf{H3}, we measured the number of user inputs across three separate tasks for each method. Here, a lower score is better: fewer inputs imply that the user is exhibiting less effort when completing the task. Paired-sample T-tests showed that participants used significantly fewer inputs when the robot used \textbf{Ours} for each task ($t(11) = 4.106, p < 0.001$, $t(11) = 5.806, p < 0.001$, $t(11) = 9.636, p < 0.001$). Figure \ref{fig:m4} shows the average magnitude of the user input over time for each task; these results further support \textbf{H3}.


Regarding \textbf{H4}, we present the subjective results from our Likert-scale survey in Figure \ref{fig:us2} (right). A one-way ANOVA analysis of the users' responses showed a significant difference in the perceived \textit{Control}, \textit{Assistance}, \textit{Prediction}, and \textit{Adaptation} that the robot exhibited when using our method ($F(69) = 11.901, p < 0.001$, $F(69) = 6.368, p < 0.005$, $F(69) = 8.794, p < 0.001$, $F(69) = 13.345, p < 0.001$). Actions chosen by \textbf{Ours} were preferable to those selected by baselines; this supports \textbf{H4}.

%% file: 6_conclusion.tex
 \section{Conclusion} \label{sec:conclusion}

In this paper, we explored the effects of communicating learned assistance back to the human operator in shared autonomy. While previous research has focused on learning the human's task and providing assistance, we instead focused on harnessing the effect of the communication. We hypothesized that humans will interact with shared autonomy systems differently when those systems communicate their learning back to the human. Using the results from online and in-person user studies, we showed that humans are more likely to intervene when the robot incorrectly predicts their intent, and release control when the robot correctly understands their task. We used the insights from these results to modify the robot's learning algorithm: under our proposed approach, the robot adjusts its model of the human's cost function to account for how communication changes the human's input patterns. Finally, we compared our approach for combining learning and communication against shared autonomy baselines that separately handle learning or communication. In a user study with 12 in-person participants across three kitchen tasks, we found that our proposed approach for combining learning and communication increased the subjective and objective performance of the human-robot team.